\documentclass[10pt,twocolumn,letterpaper]{article}

\usepackage[pagenumbers]{iccv} 

\definecolor{iccvblue}{rgb}{0.21,0.49,0.74}
\usepackage[pagebackref,breaklinks,colorlinks,allcolors=iccvblue]{hyperref}

\usepackage{arydshln}
\usepackage{color}
\usepackage{colortbl}
\usepackage{graphicx}
\usepackage{algorithm}
\usepackage{algorithmic}

\usepackage{multirow}
\usepackage{amssymb}
\usepackage{booktabs}

\usepackage{CJK}

\makeatletter
\newcommand{\thickhline}{
    \noalign {\ifnum 0=`}\fi \hrule height 1pt
    \futurelet \reserved@a \@xhline
}
\makeatother
\newcommand{\pub}[1]{\color{gray}{\scriptsize{[{#1}]}}}
\newcommand*\samethanks[1][\value{footnote}]{\footnotemark[#1]}

\title{TopV-Nav: Unlocking the Top-View Spatial Reasoning Potential of MLLM \\ for Zero-shot Object Navigation}

\author{
  Linqing Zhong$^{1}$\thanks{Equal contribution},\quad Chen Gao$^{1,2}$\samethanks,\quad Zihan Ding$^{1}$,\quad Yue Liao$^{3}$,\quad
  Huimin Ma$^{4}$,\\ Shifeng Zhang$^{5}$,\quad Xu Zhou$^{5}$,\quad Si Liu$^{1}$\thanks{Corresponding author} \\[6pt]
  {
      $^{1}$Beihang University \quad
      $^{2}$National University of Singapore
  }\\
  {
      $^{3}$MMLab, CUHK\quad
      $^{4}$USTB \quad
      $^{5}$Sangfor Technologies Inc
  }
}

\begin{document}
\maketitle
\begin{abstract}
The Zero-Shot Object Navigation (ZSON) task requires embodied agents to find a previously unseen object by navigating in unfamiliar environments. Such a goal-oriented exploration heavily relies on the ability to perceive, understand, and reason based on the spatial information of the environment. 
However, current LLM-based approaches convert visual observations to language descriptions and reason in the linguistic space, leading to the loss of spatial information. 
In this paper, we introduce TopV-Nav, an MLLM-based method that directly reasons on the top-view map with sufficient spatial information. To fully unlock the MLLM's spatial reasoning potential in top-view perspective, we propose the Adaptive Visual Prompt Generation (AVPG) method to adaptively construct semantically-rich top-view map. It enables the agent to directly utilize spatial information contained in the top-view map to conduct thorough reasoning. Besides, we design a Dynamic Map Scaling (DMS) mechanism to dynamically zoom top-view map at preferred scales, enhancing local fine-grained reasoning. Additionally, we devise a Potential Target Driven (PTD) mechanism to predict and to utilize target locations, facilitating global and human-like exploration.
Experiments on MP3D and HM3D datasets demonstrate the superiority of our TopV-Nav.
\end{abstract}
\vspace{-4mm}
\section{Introduction}
\label{sec:intro}
\vspace{-1mm}
\begin{figure}[t]
  \includegraphics[width=\columnwidth]{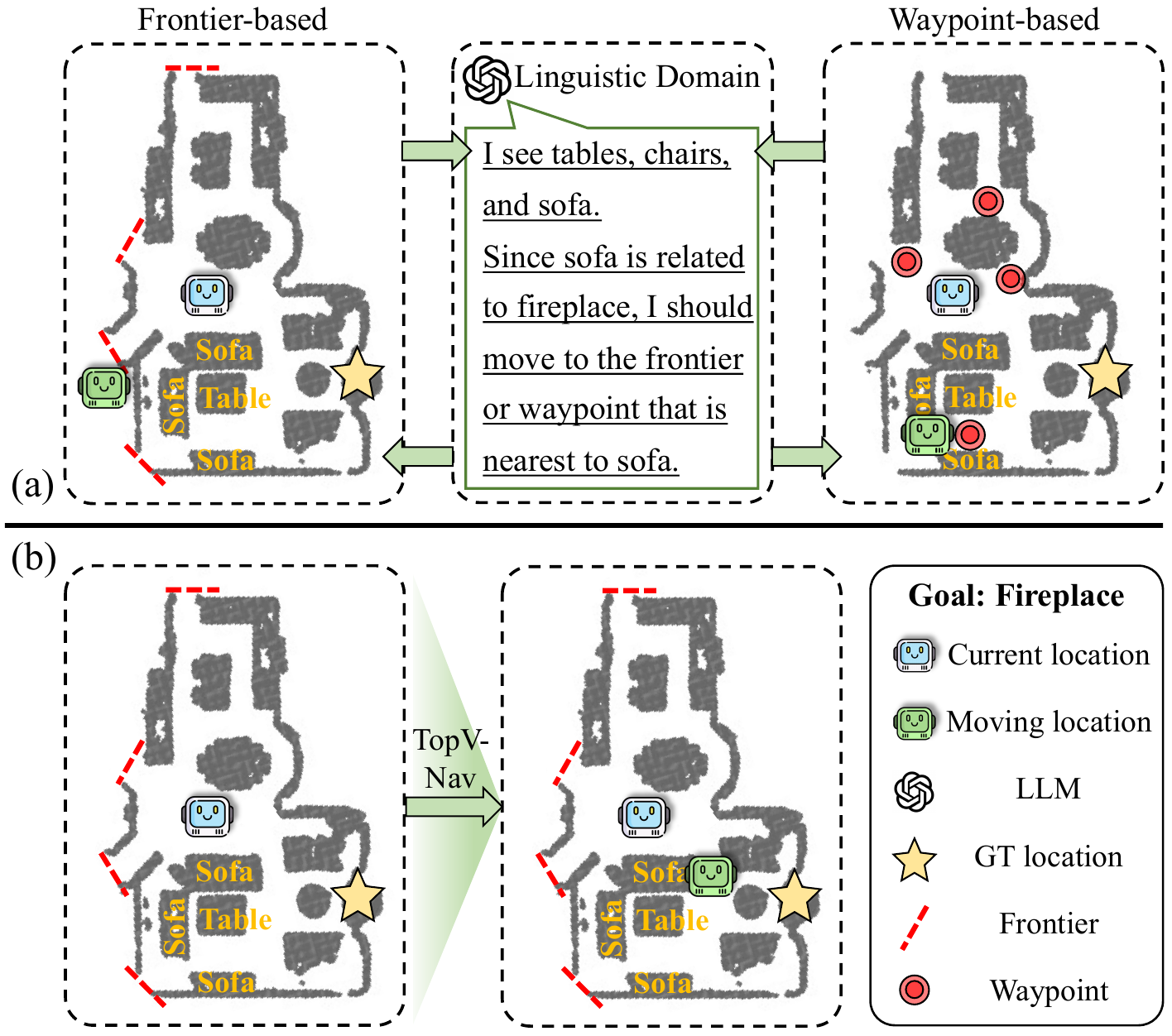}
  \vspace{-5.5mm}
  \caption{(a) Current LLM-based methods lie in two exploration paradigms, \ie, frontier-based and waypoint-based. They conduct map-to-text conversion for LLM reasoning in linguistic domain, losing the spatial information embedded in the map, \eg, room layout and spatial relation among objects. (b) TopV-Nav takes the top-view map as input and leverages MLLM to directly reason on the map image, fully utilizing the spatial information in the map.}
  \label{fig:fig1}
  \vspace{-4mm}
\end{figure}

In the realm of embodied AI, Zero-Shot Object Navigation (ZSON) is a fundamental task, requiring an agent to traverse to locate a previously-unseen object specified by category (\eg, fireplace). Such a zero-shot setting discards category-specific training and supports an open-category manner, emphasizing reasoning and exploration ability.

Recently, emerging works~\cite{voronav,esc,zhou2024navgpt} have started to integrate Large Language Models (LLMs) into ZSON agents, aiming to improve the reasoning ability by harnessing the extensive knowledge embedded in LLMs. 

As is well known, \textit{spatial information} is vital for navigation agents, as it includes essential aspects such as \textit{room layouts} and \textit{relationships among objects}. 
Typically, for encoding spatial information, navigation agents translate egocentric observations onto a structured map, \ie, top-view map or called bird's eye view map. This map serves as the core representation, facilitating essential functionalities like obstacle avoidance and path planning.

However, current LLM-based methods face notable limitations.
As shown in Fig.~\ref{fig:fig1}(a), these methods lie in two exploration paradigms, \ie, frontier-based exploration (FBE) or waypoint-based exploration.
The key issue is they need to convert the top-view map into natural language, \eg, surrounding descriptions, and use LLM to conduct reasoning in the linguistic domain. This map-to-text conversion process leads to the loss of vital spatial information such as the layout of the living room. Alternatively, if the agent knows the spatial layout of the living room and understands that the fireplace is generally positioned opposite the sofa and table, the spatial location of the fireplace can be directly inferred based on the room layout.
Therefore, considering the top-view map contains useful spatial information, and MLLMs have demonstrated capabilities in grasping spatial relationships within images in the field of image understanding~\cite{SpatialVLM,nasiriany2024pivot}, an interesting question arises \textit{\textbf{``can we leverage MLLM to reason directly on the image of top-view map to produce executable moving decisions?''}}

Besides, as shown in Fig.~\ref{fig:fig1}(a), FBE methods select a point from frontier regions to move toward exploration, restricting the LLM's action space to only the frontier boundaries.
Waypoint-based methods use the waypoint predictor to generate navigable waypoints and select a waypoint to move, restricting the LLM's action space to only a predefined set of points. 
Moreover, since the waypoint predictor is trained offline only using depth information, its predicted waypoints focus solely on traversability without semantics, also leading to weakly-semantic action space.
Both FBE and waypoint-based paradigms suffer from constrained local action spaces. Thus, the question is ``can we construct global and semantic-rich action space?''

Furthermore, when humans explore unfamiliar environments, they can use observations to infer the environment layout and predict the potential location of the target object, even target is in currently unseen areas~\cite{kessler2024human}. 
This potential target location can guide their movement decisions.
However, the action spaces of FBE and waypoint-based method are confined strictly to currently seen areas, which prevents them from possessing this capability. 
Thus, the question is ``can we leverage observations to infer the potential location of the target object to guide the current decision?''

Therefore, to address the limitations and questions mentioned above, we make multi-fold innovations.
First, we propose an insightful method called \textbf{TopV-Nav} to fully unlock the top-view spatial reasoning potential of MLLM for ZSON task. Specifically, the current LLM-driven paradigm requires the map-to-text process for LLM reasoning in linguistic space, where the converting process may lose some crucial spatial information such as objects and room layout. Instead, we propose a novel paradigm that leverages MLLM to directly reason on the top-view map, discarding the map-to-text process and maximizing the utilization of spatial information.
Second, we introduce an Adaptive Visual Prompt Generation (AVPG) method to enhance MLLM's understanding of the top-view map. AVPG adaptively generates visual prompts directly onto the map, in which various elements are spatially arranged to reflect their spatial relationships within the environment. 
Therefore, MLLM can grasp and comprehend crucial information directly from the map, facilitating effective spatial reasoning. 
For instance, in Fig.\ref{fig:fig1}(b), our method can interpret the room's layout and infer the fireplace's location.
Additionally, the moving location is predicted directly based on the top-view map, resulting in a global and semantic action space.
Third, for environments with numerous objects, the top-view map may not be able to visually represent all elements. Thus, we propose a Dynamic Map Scaling (DMS) mechanism to optionally choose a sub-region and dynamically adjust the region's scale via zooming operations. DMS further enhances the agent's local spatial reasoning and fine-grained exploration in local regions.
Last but not least, we propose a Potential Target Driven (PTD) mechanism to first predict the potential coordinate of the target object. Based on the map generated by AVPG, the predicted target coordinate can even lie in currently-unexplored areas. 
Thus, the target coordinate can guide the moving location within navigable regions, mirroring human-like predictive reasoning and exploratory behavior.
Experiments are conducted on MP3D and HM3D benchmarks, which demonstrates that our TopV-Nav achieves superior performance.
\section{Related Works}
\label{sec:related}
\begin{figure*}[t]
  \includegraphics[width=1\linewidth]{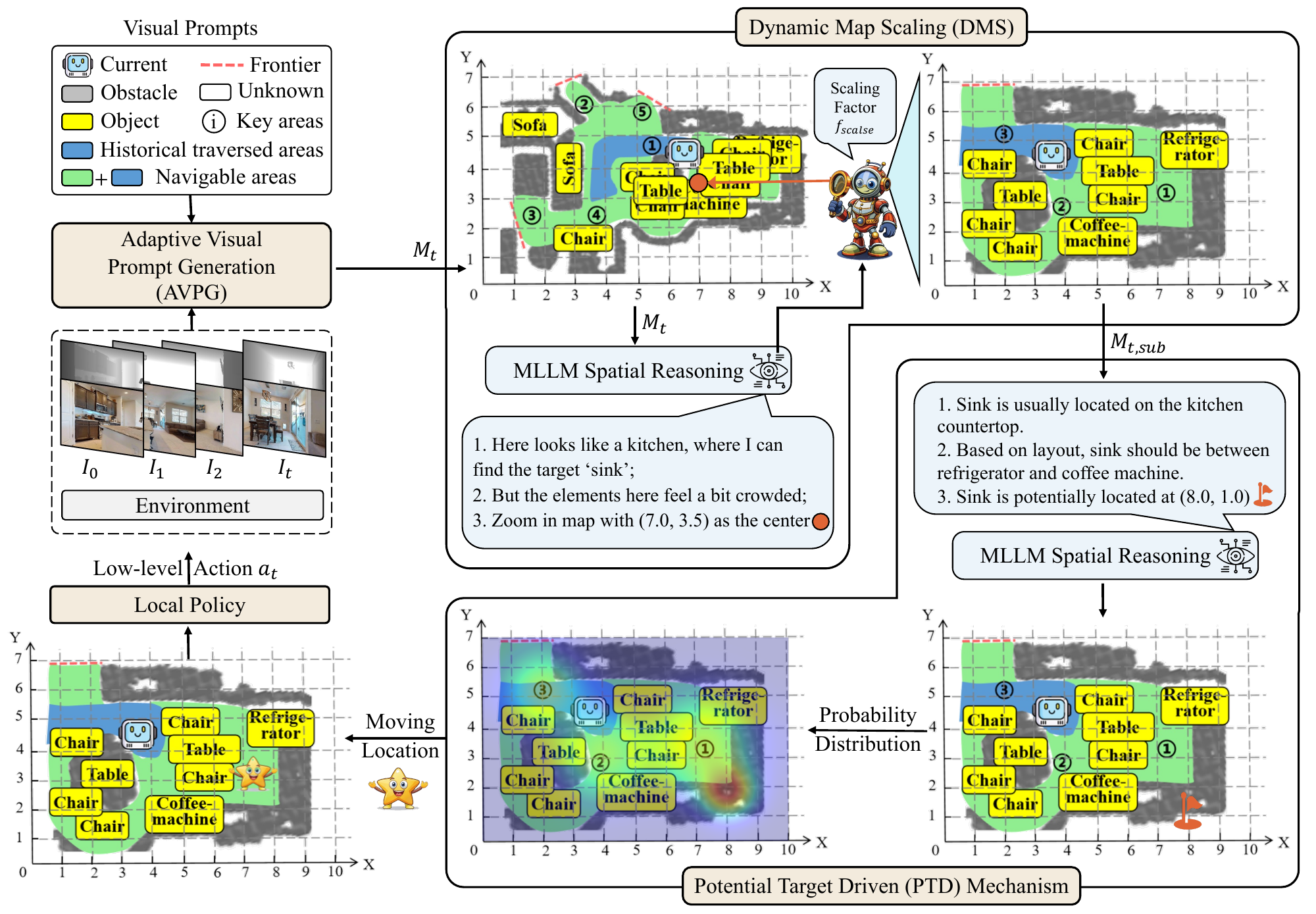}
  \vspace{-7mm}
  \caption{\textbf{Overall framework of TopV-Nav.} During navigation, the agent receives egocentric RGB-D images $I_t$ from the environment, and AVPG constructs a corresponding top-view map $M_t$. Note that visual prompts are adaptively drawn onto the map, where various elements are spatially arranged to reflect their spatial relationships.
  Subsequently, in DMS, we leverage MLLM to interpret $M_t$ and optionally select a region of interest. Then, the map is scaled according to the predicted center coordinates and dynamic scaling factor to reveal more detailed spatial information.
  Following that, in PTD, MLLM interprets the scaled map $M_{t,sub}$ to estimate the potential location of the target object and assign probability scores to key areas.
  Then, we adopt a Gaussian-based fusion strategy to obtain a value map, in which the moving location is decided accordingly. Finally, the local policy is leveraged to generate a series of low-level actions towards the moving location.
}
  \label{fig:fig2}
  \vspace{-4mm}
\end{figure*}

\subsection{Object-goal Navigation}

Object-goal navigation has been a fundamental challenge in embodied AI~\cite{Habitat-Web, PIRLNav, gao2021room, LsTDE, imaginebeforego, GMAN, neuralslam, gao2023room,peanut,chen2022reinforced, navigatingtoobjects,vsattention, hierarchical, Auxiliary,Thda,gao2023adaptive,zhao2022target}.
Early methods leverage RL to train policies, which explore visual representations~\cite{mousavian2019visual}, meta-learning~\cite{wortsman2019learning}, and semantic priors~\cite{yang2018visual,wu2018learning} to enhance performance.
Modular-based approaches~\cite{GOSE, PONI, 3DAOG} leverage perception models~\cite{he2017mask,zhang2020fusion} to construct episodic maps, based on which long-term goals are generated to guide the local policy.

To overcome closed-world assumption and achieve zero-shot object navigation (ZSON) task, EmbCLIP~\cite{embodied-clip} and \cite{ZSON} leverage the multi-modal alignment ability of CLIP~\cite{CLIP} to enable cross-domain zero-shot object navigation.
Furthermore, CoWs~\cite{CoWs} accelerates the progress of the ZSON task, where no simulation training is required, and a single model can be applied across multiple environments. 
Recent methods~\cite{voronav,esc} extract semantic information using powerful off-the-shelf detectors~\cite{glip,glipv2,groundingdino}, based on which they employ LLMs to determine the next frontier~\cite{esc} or waypoint~\cite{voronav} for exploration.
However, spatial layout information is lost during map-to-text conversion. 
To address this limitation, we investigate whether we can direct reasoning on the top-view map with MLLM, fully leveraging the complete spatial information.

\subsection{Spatial Reasoning with MLLM}
Developing spatial reasoning capabilities of MLLM has become popular recently. 
KAGI~\cite{affordance} generates a coarse robot movement trajectory as dense reward supervision.
SCAFFOLD~\cite{scaffolding} leverages scaffolding coordinates to promote vision-language coordination.
PIVOT~\cite{nasiriany2024pivot} iteratively prompts MLLM with images annotated with a visual representation of proposals and can be applied to a wide range of embodied tasks.
In the domain of vision-language navigation, AO-Planner~\cite{chen2024affordances} proposes visual affordance prompting to enable MLLM to select candidate waypoints from front-view images.
However, these works focus on exploring MLLM's spatial reasoning from egocentric perspectives, while the investigation from top-view perspective remains limited (top-view map is the core representation for robots). Although~\cite{li2024topviewrs} propose a top-view dataset, it is not designed for navigation task and lacks methods.
Our work pioneers the exploration of unlocking the top-view spatial reasoning potential of MLLM for the ZSON task.
\vspace{-1mm}
\section{Method}
\label{sec:method}
\vspace{-1mm}
In this work, we aim to investigate the question ``Can we leverage MLLM to reason directly on the top-view map to produce executable moving action for navigation agent?''. In this section, we detail the proposed MLLM-driven method, termed TopV-Nav, designed to fully unlock MLLM's top-view perception and spatial reasoning capabilities for the ZSON task. The overall framework is illustrated in Fig.~\ref{fig:fig2}. 

\vspace{-1mm}
\subsection{Problem Definition}
\vspace{-1mm}
The ZSON task requires an agent, which is randomly placed in a continuous environment as initialization, to navigate to an instance of a user-specified object category $\mathcal{G}$ in a previously unseen environment. 
At each time step $t$, the agent receives egocentric observations, which contain RGB-D images $I_t$ and its pose $p_t$. The agent is expected to adopt a low-level action $a_t$ from \texttt{move\_forward}, \texttt{turn\_left}, \texttt{turn\_right}, \texttt{look\_up}, \texttt{look\_down} and \texttt{stop}. The task is considered successful if the agent stops within a distance threshold from the target and the target is visible in the egocentric observation. 

\subsection{Adaptive Visual Prompt Generation}
\label{sec:AVPG}
To construct a top-view map that enables MLLM to effectively understand and utilize spatial information for navigation decision, we propose the Adaptive Visual Prompt Generation (AVPG) module.

Intuitively, a comprehensible top-view map for MLLM to conduct navigation should contain elements: current location, historical traversed areas, obstacles, frontiers, objects' location/category, \etc.
Therefore, we build map utilizing visual prompts to reflect these elements.
As shown in the top left corner of Fig.~\ref{fig:fig2}, we adopt different colors and text to denote different elements on the map.

Technically, to construct the top-view map $M_t$ at each time step $t$, we first transform the agent's egocentric RGB-D images into 3D point clouds utilizing the agent's pose $p_t$. 
Then, we classify points near the floor as part of the navigable areas, while points exceeding a height threshold are identified as obstacles. Next, we project these points onto $M_t$. 
Moreover, we employ a detector to identify objects from the agent’s egocentric RGB images and project them onto $M_t$. Subsequently, text-boxes as visual prompts are drawn on $M_t$, indicating each object's location and category. 
The MLLM is thus empowered to precisely recognize key entities by observing the top-view map.

\vspace{0.5mm}
\noindent\textbf{Key Area Markers Generation.} 
To make MLLM better interpret $M_t$, we generate markers as visual prompts on the map to refer key areas that contain rich semantics. 
Specifically, we apply a density-based spatial clustering algorithm to group both frontiers and objects on the map, producing markers $\{m_1,m _2,\dots,m_{N_{m}}\}$ to represent $N_{m}$ key areas. Note that $N_{m}$ is adaptively changed during navigation.

Technically, we first identify each frontier that separates the explored and unexplored areas and obtain its midpoint $f$. These midpoints $\{f_1,f_2,\dots,f_{N_{f}}\}$ are considered as candidate points for clustering, where $N_{f}$ is the number of frontiers. $\mathcal{O}$ denotes the set of detected objects.
Then, we randomly sample a point $p_i \in \mathcal{O}\cup\mathcal{F}$ which is not yet clustered. Centering around $p_i$, we consider its $\epsilon$-neighborhood and calculate the number of its neighboring points $N_{\epsilon}(p_i)$, which is formulated as:
\begin{equation}
N_{\epsilon}(p_i) = \{p_j \mid \lVert p_i - p_j \rVert_2 \leq \epsilon \},
\end{equation}
where $\epsilon$ denotes the maximum distance between two points in the same neighborhood. $p_i$ is classified as an element of $i$-th key area $A_i$ if the number of its neighboring points $N(p_i)$ satisfies $N(p_i) \geq min\_pts$. Note that $min\_pts$ indicates the minimum number of points required to form key area $A_i$.  Otherwise, $p_i$ is considered as an outlier and will be removed from the calculation.

If the center points of two key areas $A_i$ and $A_j$ fall within each other's $\epsilon$-neighborhood, we merge them to form a new area. The computation is iteratively conduct until all points are assigned to a key area or seen as outliers. Eventually, each key area marker $m_i$ is defined as the centroids of each key area $A_i$, which is obtained via:
\begin{equation}
m_{i} = \frac{1}{\mid A_i\mid}\sum_{p_j \in A_i}{p_j}.
\end{equation}
These key area markers are utilized to indicate the semantically significant regions on the top-view map $M_t$, facilitating MLLM to perform spatial reasoning. 
We further overlay a coordinate system and grid lines on $M_t$ to precise spatial references, enabling MLLM to accurately determine spatial position/relation among different elements.

\begin{figure}[t]
  \includegraphics[width=\columnwidth]{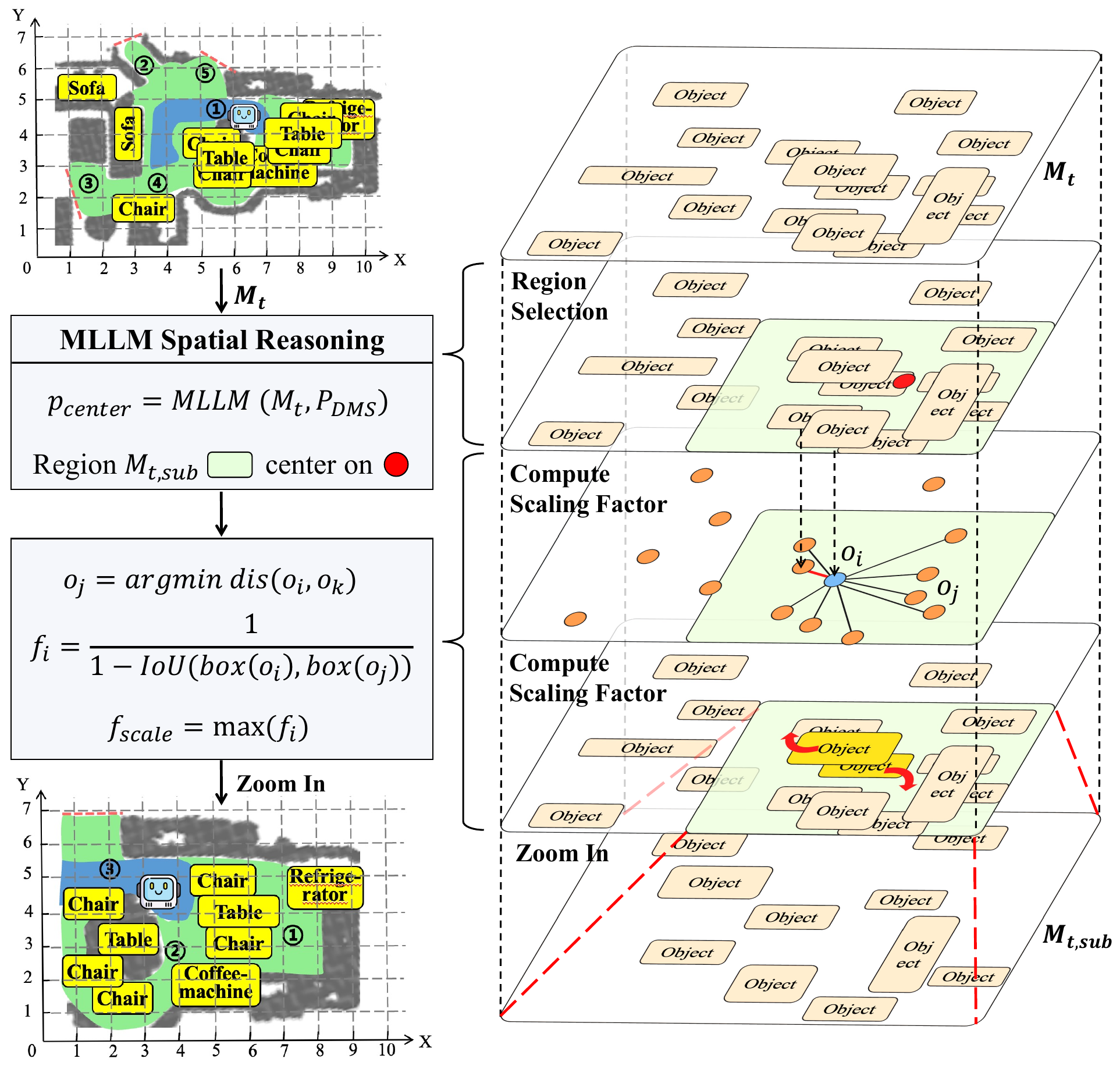}
  \vspace{-6mm}
  \caption{Illustration of the DMS mechanism.}
  \label{fig:fig3}
  \vspace{-4mm}
\end{figure}

\subsection{Dynamic Map Scaling}
On $M_t$, visual prompts may overlap with each other (\eg, between text-boxes as shown in Fig.~\ref{fig:fig3}), hindering the MLLM from capturing all clues. 
Instructively, when humans zoom in the GoogleMaps on phones using two-finger gestures, the previously overlapping elements will be fully expanded accordingly, thereby revealing more spatial details. Inspired by such mechanism, we devise the DMS.

\vspace{0.5mm}
\noindent\textbf{Region Selection.}
Recall that a coordinate system is drawn on $M_t$ as positional references, allowing MLLM to identify precise spatial positions.
As shown in Fig.~\ref{fig:fig3}, we prompt MLLM to interpret $M_t$ and determine which region on $M_t$ needs a closer observation. Concretely, we adopt MLLM to directly predict the center coordinate of the region via $p_{center} = \operatorname*{MLLM}(M_t, P_{DMS})$, where $P_{DMS}$ denotes the textural prompt. Then, we extract the maximal edge-aligned rectangle region $M_{t,sub}$ with $p_{center}$ as the center. 

\vspace{0.5mm}
\noindent\textbf{Compute Scaling Factor.}
After determining the region, a significant challenge is obtaining an appropriate scaling factor $f_{scale}$ to enlarge $M_{t,sub}$, allowing the map to be scaled up so that all visual elements are fully displayed without being obstructed. 
Thus, we devise a layout-aware strategy to dynamically calculate the appropriate scaling factor.

Concretely, $\mathcal{O}_{sub}=\{o_1,o_2,\dots,o_{N_{obj}}\}$ is the set of detected object in $M_{t,sub}$, where $o_i$ is the coordinates of the text-box's center of $i$-th object. As shown in Fig.~\ref{fig:fig3}, for each $o_i$, we identify the nearest neighboring object point $o_j$ through computing the Euclidean distance as follows:
\begin{equation}o_j = \operatorname*{arg\,min}_{o_k\in\mathcal{O}_{sub}\backslash\{o_i\}}\lVert o_i - o_k \rVert_2.
\end{equation}
Then, we compute the Intersection over Union (IoU) of the text-boxes of $o_i$ and $o_j$ to quantify the occlusion degree. 
Intuitively, a higher IoU between text-boxes indicates $M_{t,sub}$ should be further scaled to prevent text-boxes from overlapping. We define the scaling factor $f_{i}$ for $p_i$ as following:
\begin{equation}
f_{i} = \frac{1}{1-\operatorname*{IoU}(box(o_i), box(o_j))}.
\end{equation}
Thus $f_i$ represents the ideal map scaling factor to avoid occlusion between objects $o_i$ and $o_j$. Following this, we iteratively examine each point within $\mathcal{O}_{sub}$ and compute its corresponding scaling factor. The scaling factor $f_{scale}$ for the entire map $M_{t,sub}$ is determined as the maximum value across all individual scaling factors, which is formalized as:
\begin{equation}f_{scale} = \operatorname*{max}_{o_i\in \mathcal{O}_{sub}}(f_i).
\end{equation}
This strategy ensures that $f_{scale}$ is adaptively adjusted. Eventually, given the final scaling factor $f_{scale}$, we zoom in $M_{t,sub}$ accordingly and regenerate the visual prompt by AVPG. Note that we keep $M_{t,sub}$ standardized through centering on $p_{center}$ and cutting off regions that extend beyond the resolution limitation, which ensures the consistency of map scale. 
For simplicity, we uniformly denote scaled and unscaled top-view maps as $M_t$ in the following sections.

\subsection{Potential Target Driven Mechanism}
\label{sec:PTD}
When searching for a target object in an unknown environment, humans can infer the potential location of the target based on known observations, even if this location lies in unexplored areas. This estimated location can then guide their current movement decisions. Inspired from this, we propose the Potential Target Driven (PTD) mechanism.

\begin{figure}[t]
  \includegraphics[width=\columnwidth]{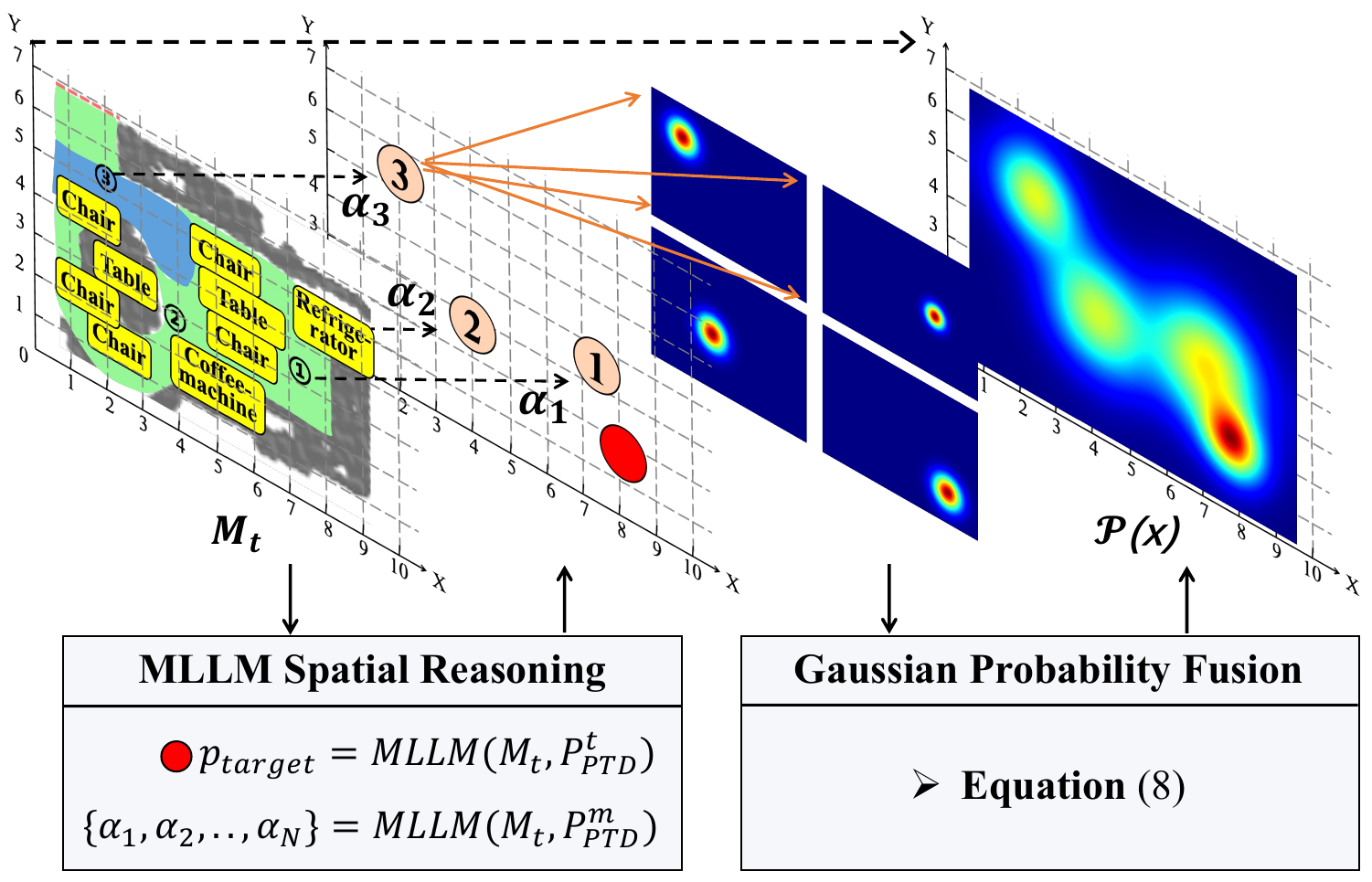}
  \vspace{-4mm}
  \caption{Illustration of the PTD mechanism.}
  \label{fig:fig4}
  \vspace{-2mm}
\end{figure}

\vspace{0.5mm}
\noindent\textbf{Potential Target Prediction.} As shown in Fig.~\ref{fig:fig4}, given the top-view map $M_t$, we prompt the MLLM to predict the potential location $p_{target}$ of target object via spatial reasoning:
\begin{equation}p_{target} = \operatorname*{MLLM}(M_t, P^{t}_{PTD}),
\end{equation}
where $P^{t}_{PTD}$ denotes the textual prompts.

\vspace{0.5mm}
\noindent\textbf{Probability Fusion.} To leverage $p_{target}$ to guide navigation, we adopt Gaussian-based probability fusion strategy. Specifically, we also prompt the MLLM to assign probability scores $\{\alpha_1, \alpha_2, ..., \alpha_{N_{mk}}\}$ for $N_{mk}$ key area markers:
\begin{equation}
\{\alpha_1, \alpha_2, ..., \alpha_{N_{mk}}\} = \operatorname*{MLLM}(M_t, P^{m}_{PTD}),
\end{equation}
where $P^{m}_{PTD}$ represents the text prompt querying the MLLM to assign probability scores to key area markers. Note that these markers are clustered within the entire navigable region, thus resulting in global action space.

Subsequently, we construct a two-dimensional Gaussian distribution centered around each key area marker $\{m_1, m_2, ..., m_{N_{mk}}\}$ and the potential target location $p_{target}$. 
Then, through applying the superposition of these distributions, we obtain a fused Gaussian probability distribution $\mathcal{P}(x)$ map, which indicates the likelihood of the target object being present at each location $x$, as shown in Fig.~\ref{fig:fig4}. Such process can be formulated as follows:
\begin{equation}
\small
\mathcal{P}(x) = \sum_{i=1}^{N_{mk}}\alpha_i \cdot \mathcal{N}(x|m_i, \sigma_i^2) + \beta \cdot \mathcal{N}(x|p_{target}, \sigma_{target}^2),
\end{equation}
where $\mathcal{N}(x|\mu, \sigma^2)$ represents a normalized 2D Gaussian distribution. We utilize scores $\{\alpha_1, \alpha_2, ..., \alpha_{N_{mk}}\}$ and a hyper-parameter $\beta$ as peak values of distributions.
$\sigma_i$ and $\sigma_{target}$ are dynamically calculated to ensure that the probability map can gradually decrease to $0.1$ at its farthest marker.
Eventually, we take the location with the highest probability as the agent's actual moving location.

\definecolor{mygray}{gray}{.9}
\begin{table*}
   \begin{center}
      \resizebox{0.87\linewidth}{!}{
         \centering
\resizebox{1\textwidth}{!}{
\setlength\tabcolsep{3.5pt} %
\renewcommand\arraystretch{1.25} %
    \begin{tabular}{rl||c|c|c|cc|cc}
    \hline \thickhline
    \rowcolor[gray]{0.96}
    ~ & & & & & \multicolumn{2}{c|}{MP3D} & \multicolumn{2}{c}{HM3D} \\
    \cline{6-9}
    \rowcolor[gray]{0.96}
    \multicolumn{2}{c||}{\multirow{-2}{*}{Methods}}  & \multirow{-2}{*}{Zero-Shot}  & \multirow{-2}{*}{{Training-Free}} & \multirow{-2}{*}{{Reasoning Domain}} & \texttt{SR}\!~$\uparrow$ & \texttt{SPL}\!~$\uparrow$  & \texttt{SR}\!~$\uparrow$  & \texttt{SPL}\!~$\uparrow$   \\
    \hline
    \hline
    SemEXP$\!$~\cite{GOSE} & $\!\!$\pub{NeurIPS2020} & $\times$ & $\times$ & Latent Map & 36.0 & 14.4 & - & - \\
    PONI$\!$~\cite{PONI} & $\!\!$\pub{CVPR2022} & $\times$ & $\times$ & Latent Map & 31.8 & 12.1 & - & - \\
    ProcTHOR$\!$~\cite{procthor} & $\!\!$\pub{NeurIPS2022} & $\times$ & $\times$ & CLIP Embeddings & - & - & 54.4 & 31.8 \\
    \hline
    ProcTHOR-ZS$\!$~\cite{procthor} & $\!\!$\pub{NeurIPS2022} & \checkmark & $\times$ & CLIP Embeddings & - & - & 13.2 & 7.7 \\
    ZSON$\!$~\cite{ZSON} & $\!\!$\pub{NeurIPS2022} & \checkmark & $\times$ & CLIP Embeddings & 15.3 & 4.8 & 25.5 & 12.6 \\
    PSL$\!$~\cite{sun2024prioritized} &$\!\!$\pub{ECCV2024} & \checkmark & $\times$ & CLIP Embeddings & - & - & 42.4 & 19.2 \\
    Pixel-Nav$\!$~\cite{cai2024bridging} & $\!\!$\pub{ICRA2024} & \checkmark & $\times$ & Linguistic & - & - & 37.9 & 20.5 \\
    SGM$\!$~\cite{imaginebeforego} & $\!\!$\pub{CVPR2024} & \checkmark & $\times$ & Linguistic & 37.7 & 14.7 & 60.2 & 30.8 \\
    ImagineNav & $\!\!$\pub{ICLR2025} & \checkmark & $\times$ & Linguistic & - & - & 53.0 & 23.8 \\
    \hline
    CoW$\!$~\cite{CoWs} & $\!\!$\pub{CVPR2023} & \checkmark & \checkmark & CLIP Embeddings & 7.4 & 3.7 & - & - \\
    ESC$\!$~\cite{esc} & $\!\!$\pub{ICML2023} & \checkmark & \checkmark & Linguistic & 28.7 & 14.2 & 39.2 & 22.3 \\
    VoroNav$\!$~\cite{voronav} & $\!\!$\pub{ICML2024} & \checkmark & \checkmark & Linguistic & - & - & 42.0 & 26.0 \\
    \textbf{{TopV-Nav}} & (Ours) & \checkmark & \checkmark & Top-view Map & \textbf{35.2} & \textbf{16.4} & \textbf{52.0} & \textbf{28.6} \\
    \hline
\end{tabular}
}

      }
   \end{center}
   \vspace{-5mm}
   \caption{\textbf{Main Comparisons}. Our TopV-Nav significantly boosts the performance in terms of the key metrics on benchmarks.}
   \label{tab:sota}
   \vspace{-4mm}
\end{table*}

\subsection{Local Policy}
\label{sec:local policy}
Following previous works~\cite{yang2023iplanner, liang2024mtg}, once PTD produces agent's actual moving location, we then leverage a local controller to conduct a series of low-level actions such as \texttt{move\_forward} and \texttt{turn\_left}. These low-level actions make the agent move toward the actual moving location gradually.
If the target object is discovered during this process, the agent subsequently navigates to it and eventually calls the \texttt{STOP} action to accomplish the task. Note that exceeding the action step limit will trigger a forced stop.

\definecolor{mygray}{gray}{.9}
\section{Experiments}
\vspace{-1mm}

\subsection{Experimental Setup}
\vspace{-1mm}

\noindent\textbf{Dataset.}
We conduct experiments on two representative object navigation benchmarks, \ie, Matterport3d (MP3D)~\cite{chang2017matterport3d} and Habitat-Matterport3D (HM3D)~\cite{yadav2023habitat}. MP3D provides $2,195$ episodes in $11$ indoor scenes for validation, with $21$ categories of object goal. HM3D offers high-fidelity reconstructions of $20$ entire buildings, and incorporates $2,000$ validation episodes for the task with $6$ categories.

\vspace{0.5mm}
\noindent\textbf{Evaluation Metrics.}
Representative metrics are adopted, \ie, Success Rate (SR) and Success weighted by Path Length (SPL). SR is defined as the proportion of episodes where the agent’s distance to the target object is less than $1$m after \texttt{STOP}. SPL considers navigation efficiency by accounting for both success and the ratio of the shortest path length to the actual path length taken by the agent.

\vspace{0.5mm}
\noindent\textbf{Implementation Details.}
We construct $M_t$ with a resolution of $1,000 \times 1,000$, where each meter in the real world corresponds to $20$ pixels, ensuring $M_t$ is large enough to cover the unknown environment. We utilize $\epsilon=1.3$m and $min\_pts=2$ in the clustering algorithm to generate key area markers. In the DMS module, we impose an upper bound of $5$ on $f_{scale}$ to mitigate excessive map scaling resulting from complete text-box overlap. We set the peak value of the Gaussian probability distribution centered around the potential target location as $\beta=0.5$ in PTD. 
We adopt Qwen2.5-VL-7B~\cite{bai2025qwen2} as our MLLM. Comparisons about using different MLLM are shown in Sec.~\ref{sec:exp-abs}.

\vspace{-1mm}
\subsection{Comparison with Previous Methods}
\vspace{-1mm}

We compare our TopV-Nav with prior state-of-the-art methods. 
Compared to the methods within the training-free setting, ESC leverages LLM's commonsense knowledge such as object co-occurrence to guide navigation, while VoroNav utilizes LLMs to select the agent's optimal traversal path among path descriptions. Both of them rely on LLMs for reasoning in linguistic space. 
Essentially, we investigate the potential of MLLM's top-view spatial reasoning against vanilla LLM's reasoning in the linguistic domain.

As shown in Tab.~\ref{tab:sota}, on both MP3D and HM3D datasets, our proposed TopV-Nav significantly outperforms priors works. Specifically, our method outperforms ESC on MP3D by improving SR and SPL by $6.5\%$ and $2.2\%$. On the HM3D dataset, TopV-Nav increases the SR from $42.0\%$ to $52.0\%$ while the SPL is simultaneously raised from $26.0\%$ to $28.6\%$. Consistently, recall that these LLM-based works convert visual information into textual descriptions for LLM reasoning. 
However, this transformation results in the loss of spatial cues, which is crucial for navigation. Through leveraging MLLM to directly reason on the top-view map, our TopV-Nav fully preserves spatial information, which serves as navigation guidance for agent's decision-making process. The performance improvement demonstrates the superiority of our proposed method.

\subsection{Ablation Studies}
\label{sec:exp-abs}

We conduct extensive ablation studies(shown in Tab.~\ref{tab:ablation},~\ref{tab:avpg},~\ref{tab:mllms}, and~\ref{tab:ptd}). 
Due to the cost, we sample a subset of HM3D for ablations, which cover all validation scenes and target object categories, ensuring representativeness and fairness.

\vspace{0.5mm}
\noindent\textbf{Adaptive Visual Prompt Generation.}
As shown in Tab.~\ref{tab:ablation}, we utilize LLM for reasoning only in linguistic space (\ie, without map input) as the Baseline.
Note that experiment ``\#1'' introduces the AVPG module to the baseline, leveraging MLLM to take the top-view map generated by AVPG as the input. As the results show, compared to baseline, experiment ``\#1'' boosts SR from $45.0\%$ to $49.0\%$ and improves SPL from $25.44\%$ to $28.07\%$. 
It reveals that compared with reasoning only in the linguistic domain, the MLLM can offer spatial navigation guidance through conducting spatial reasoning on the top-view map generated by AVPG.

\begin{table}
   \begin{center}
      \resizebox{0.9\linewidth}{!}{
         \centering
\resizebox{1\textwidth}{!}{
\renewcommand\arraystretch{1.25} %
    \begin{tabular}{c||c|c|c|cc}
    \hline \thickhline
    \rowcolor[gray]{0.96}
    ~ & & & & \multicolumn{2}{c}{HM3D}  \\
    \cline{5-6}
    \rowcolor[gray]{0.96}
    \multirow{-2}{*}{Name} & \multirow{-2}{*}{AVPG} & \multirow{-2}{*}{DMS} & \multirow{-2}{*}{PTD}   & \texttt{SR}\!~$\uparrow$  & \texttt{SPL}\!~$\uparrow$   \\
    \hline
    \hline
    LLM-based  &  &  & \\
    Baseline  &  &  &  &\multirow{-2}{*}{45.0} & \multirow{-2}{*}{25.44} \\
    \hline
    \#1  & \checkmark &  &  & 49.0 & 28.07 \\
    \#2  & \checkmark & \checkmark & & 50.0 & 27.16 \\
    \#3 & \checkmark  & \checkmark & \checkmark & \textbf{52.0} & \textbf{28.73} \\
    \hline
\end{tabular}
}
      }
   \end{center}
   \vspace{-4mm}
   \caption{\textbf{Main Ablations}. The performance is improved with the continuous addition of proposed methods, verifying the effectiveness of each component.}
   \label{tab:ablation}
   \vspace{-4mm}
\end{table}
\vspace{0.5mm}

\noindent\textbf{Visual Prompt Components.}
We conduct ablation studies to investigate the effects of different visual prompts, which is shown in Tab.~\ref{tab:avpg}.
The comparison between ``Full Prompt” and the other lines demonstrates the effectiveness of incorporating these visual prompts. Moreover, the ablation of different visual prompts provides valuable insights. Notably, removing text-boxes that represent objects and the coordinate grid leads to a significant performance drop. Intuitively, the text-boxes enrich the map with semantic information, while the coordinate grid offers spatial references for MLLM, both of which are crucial for navigation.

\vspace{0.5mm}
\noindent\textbf{Dynamic Map Scaling.}
Through comparing ``\#1'' and ``\#2'' in Tab.\ref{tab:ablation}, we observe that the DMS module promotes the SR from $49.0\%$ to $50.0\%$, which demonstrates the effectiveness of the DMS module.
Also, a slight reduction $0.91\%$ in SPL reveals that more fine-grained local exploration involves a trade-off, \ie, longer trajectory. Intuitively, scaling the local region increases the likelihood of discovering the target but may also lead to more exploration.

\vspace{0.5mm}
\noindent\textbf{Potential Target Driven Mechanism.}
As shown in ``\#3'' in Tab.\ref{tab:ablation}, compared to ``\#2'', SR is raised up from $50.0\%$ to $52.0\%$ and SPL also gains $1.57\%$ absolute increment. 
This improvement on both metrics validates the PTD mechanism and also confirms that human-like predictive reasoning leads to improvement of navigation performance.

In Tab.~\ref{tab:ptd}, we investigate the effects of fusion policies in PTD. Specifically, directly selecting the moving location with the highest probability score without fusion (denoted as ``Max'') and constructing Gaussian-based fusion map for selecting the moving location (noted as ``Gaussian'').
As the results show, with fusion policy, the results outperform just selecting the candidate point with the highest probability. Moreover, we conduct an analysis on $\beta$, \ie, the probability scores assigned to the potential target location. We observe that setting $\beta=0.5$ achieves the best performance.

\begin{table}
   \begin{center}
      \resizebox{0.6\linewidth}{!}{
         \centering
\resizebox{1\textwidth}{!}{
\renewcommand\arraystretch{1.25} %
    \begin{tabular}{c||cc}
    \hline \thickhline
    \rowcolor[gray]{0.96}
    ~ &\multicolumn{2}{c}{HM3D}  \\
    \cline{2-3}
    \rowcolor[gray]{0.96}
    \multirow{-2}{*}{AVPG} & \texttt{SR}\!~$\uparrow$  & \texttt{SPL}\!~$\uparrow$   \\
    \hline
    \hline
    Full Prompt & \textbf{52.0}  & \textbf{28.73} \\
    \hline
    w/o History & 51.0 & 28.51\\
    w/o Obstacle & 49.0 & 26.88 \\
    w/o Text-boxes & 45.0 & 26.16 \\
    w/o Coordinate & 46.0 & 26.08 \\
    \hline
\end{tabular}
}
      }
   \end{center}
   \vspace{-4mm}
   \caption{\textbf{Ablations}. We examine the visual prompts' effectiveness.}
   \label{tab:avpg}
   \vspace{-1mm}
\end{table}

\begin{table}
   \begin{center}
      \resizebox{0.8\linewidth}{!}{
         \centering
\resizebox{1\textwidth}{!}{
\renewcommand\arraystretch{1.25} %
\begin{tabular}{c||c|cc}
    \hline \thickhline
    \rowcolor[gray]{0.96}
    ~ & ~ &\multicolumn{2}{c}{HM3D}  \\
    \cline{3-4}
    \rowcolor[gray]{0.96}
    \multirow{-2}{*}{PTD} & \multirow{-2}{*}{$\beta$} & \texttt{SR}~$\uparrow$ & \texttt{SPL}~$\uparrow$ \\
    \hline
    \hline
    \multirow{3}{*}{w/o Fusion (Max)} & 0.4 & 48.0 & 26.59 \\
    & 0.5 & 50.0 & 27.29\\
    & 0.6 & 49.0 & 26.28\\
    \hline
    \multirow{3}{*}{w/ Fusion (Gaussian)} & 0.4 & 52.0 & 28.15 \\
    & 0.5 & \textbf{52.0} & \textbf{28.73} \\
    & 0.6 & 51.0 & 27.28\\
    \hline
\end{tabular}
}
      }
   \end{center}
   \vspace{-4mm}
   \caption{\textbf{Ablations}. We investigate the effects of fusion policy and hyper-parameter $\beta$ utilized in the PTD.}
   \label{tab:ptd}
   \vspace{-1mm}
\end{table}

\begin{table}
   \begin{center}
      \resizebox{0.85\linewidth}{!}{
         \centering
\resizebox{1\textwidth}{!}{
\renewcommand\arraystretch{1.25} %
    \begin{tabular}{c||c|cc}
    \hline \thickhline
    \rowcolor[gray]{0.96}
    ~ & & \multicolumn{2}{c}{HM3D}  \\
    \cline{3-4}
    \rowcolor[gray]{0.96}
    \multirow{-2}{*}{MLLM Name} & \multirow{-2}{*}{Backbone LLM} & \texttt{SR}\!~$\uparrow$  & \texttt{SPL}\!~$\uparrow$   \\
    \hline
    \hline
    LLaVA-NeXT & LLama-3-8B  & 50.0 & 27.66\\
    Qwen-2.5-VL & Qwen2.5-7B & 52.0 & 28.73 \\
    GPT-4o & - & \textbf{53.0} & \textbf{29.78} \\
    \hline
\end{tabular}
}
      }
   \end{center}
   \vspace{-4mm}
   \caption{\textbf{Ablations}. We examine the effects by adopting different open-/closed source MLLMs.}
   \label{tab:mllms}
   \vspace{-3mm}
\end{table}

\begin{figure*}[t]
  \includegraphics[width=1\linewidth]{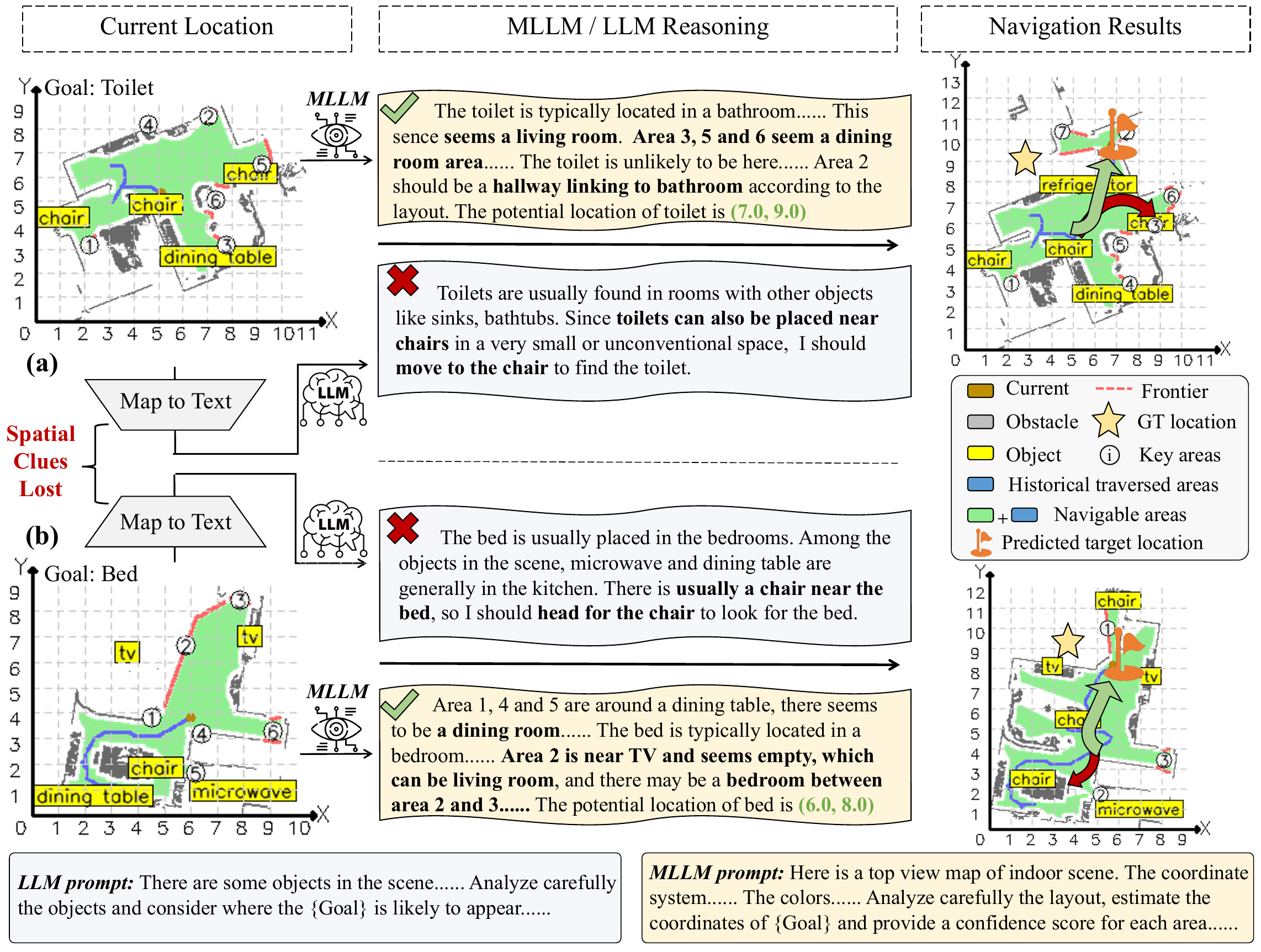}
  \vspace{-5mm}
  \caption{Qualitative comparisons of navigation decisions between TopV-Nav and LLM-based baseline. Best viewed in color.}
  \label{fig:visualization}
  \vspace{-3mm}
\end{figure*}

\vspace{1mm}
\noindent\textbf{Analysis on Different MLLMs.}
We also conduct a comparative analysis for different MLLMs. As shown in Tab.~\ref{tab:mllms}, the GPT-4o achieves the highest performance in both SR and SPL. However, due to its closed-source nature and high costs, we leverage open-source Qwen-2.5-VL as our MLLM to construct TopV-Nav in this work. Ideally, with the advancement of MLLM technology in the future, our method's performance can be directly improved by replacing the model with a better MLLM.

\subsection{Qualitative Analysis}
\vspace{1mm}

To give a more intuitive view, we visualize the TopV-Nav's navigation process (shown in Fig.~\ref{fig:visualization}), also comparing it with the LLM-based baseline method that only adopts LLM reasoning in the linguistic domain.

In Fig.~\ref{fig:visualization}(a), the agent is tasked to search for a ``toilet'' in an indoor environment. In its current location, the representative objects of living-room and dining room such as ``chairs'' are located in this area. The MLLM recognizes the living room and dining room through observing the scene layout. Subsequently, it performs spatial reasoning and infer that there should be a hallway linking to the bathroom in area 2. Due to toilet being present in bathroom, it estimates the potential location of ``toilet'' to be ``(7.0, 9.0)''. Turing to LLM, since it necessitates converting visual information to textual description, the vital spatial clues are lost. Therefore, it only relies on object co-occurrence for reasoning and naturally leads the agent to the wrong direction.

In Fig.~\ref{fig:visualization}(b), the target object is ``bed''. In LLM's reasoning, due to the lack of spatial information, LLM only considers a chair often appears near a bed and assumes it as the agent's moving location. However, from the top-view map, the chair is clearly located in the dining room, where the agent cannot find the bed. In contrast, reasoning based on the full spatial layout, MLLM identifies area 2 as a living room even though it has not been fully explored. Moreover, MLLM infers that there may be a bedroom between areas 2 and 3, where the agent could find the bed. By setting ``(6.0, 8.0)'' as the potential target location, agent is guided to explore and eventually discovers the bedroom.
\section{Conclusion}
\label{sec:conclusion}
\vspace{-1.5mm}

In this paper, we tackle the Zero-Shot Object Navigation (ZSON) task, where spatial information plays a critical role in such a goal-oriented exploration task. However, previous LLM-based methods transfer the top-view map to language descriptions, conducting reasoning in the linguistic domain. This transformation process loses spatial information such as object and room layout. Therefore, we aim to study how we can directly adopt the top-view map for reasoning by using MLLM's image understanding ability. Specifically, we propose several insightful methods to fully unlock the top-view spatial reasoning potential of MLLM. The proposed Adaptive Visual Prompt Generation (AVPG) method draws a semantically-rich map with visual prompts. The Dynamic Map Scaling (DMS) mechanism adjusts the map scale dynamically interpreting layout and decision-making. The Potential Target Driven (PTD) mechanism imitates human behavior to predict the target's potential location to guide the current action. Experiments on MP3D and HM3D benchmarks demonstrate the superiority of our method.

{
    \small
    \bibliographystyle{ieeenat_fullname}
    \bibliography{main}
}

\end{document}